\documentclass[10pt,twocolumn,letterpaper]{article}

\usepackage{iccv}
\usepackage{times}
\usepackage{epsfig}
\usepackage{graphicx}
\usepackage{amsmath}
\usepackage{amssymb}

\usepackage[sort&compress,square,comma,numbers]{natbib}
\makeatletter
\renewcommand\bibsection%
{
  \section*{\refname
    \@mkboth{\MakeUppercase{\refname}}{\MakeUppercase{\refname}}}
}
\makeatother

\usepackage{pgfplots}
\usepackage{pgfplotstable}
\usepackage{filecontents,pgfplots}
\usepackage{tikz}
\usetikzlibrary{arrows,calc,shapes,positioning,matrix,arrows,decorations.pathmorphing}

\newcommand{\EE}{\mathcal{E}}
\newcommand{\DD}{\mathcal{D}}
\newcommand{\RR}{\mathcal{R}}

\DeclareMathOperator*{\argmax}{argmax}
\DeclareMathOperator*{\argmin}{argmin}

\usepackage[pagebackref=true,breaklinks=true,letterpaper=true,colorlinks,bookmarks=false]{hyperref}

\iccvfinalcopy 

\def\iccvPaperID{1552} 

\ificcvfinal\fi
\begin{document}

\title{Self-Supervised Representation Learning via Neighborhood-Relational Encoding}


\author{Mohammad Sabokrou\\
{\normalsize Institute for Research in Fundamental Sciences}\\
{\tt\small sabokro@ipm.ac.ir}
 \and
Mohammad Khalooei \\
{\normalsize Amirkabir University of Tech.}\\
{\tt\small khalooei@aut.ac.ir}
\and
Ehsan Adeli \\
{\normalsize  Stanford University}\\
{\tt\small eadeli@cs.stanford.edu}
}

\maketitle

\begin{abstract}
In this paper, we propose a novel self-supervised representation learning by taking advantage of a neighborhood-relational encoding (NRE) among the training data. Conventional unsupervised learning  methods only focused on training deep networks to understand the primitive characteristics of the visual data, mainly to be able to reconstruct the data from a latent space. They often neglected the relation among the samples, which can serve as an important metric for self-supervision. Different from the previous work, NRE aims at preserving the local neighborhood structure on the data manifold. Therefore, it is less sensitive to outliers. We integrate our NRE component with an encoder-decoder structure for learning to represent samples considering their local neighborhood information. Such discriminative and unsupervised representation learning scheme is adaptable to different computer vision tasks due to its independence from intense annotation requirements. We evaluate our proposed method for different tasks, including classification, detection, and segmentation based on the learned latent representations. In addition, we adopt the auto-encoding capability of our proposed method for applications like defense against adversarial example attacks and video anomaly detection. Results confirm the performance of our method is better or at least comparable with the state-of-the-art for each specific application, but with a generic and self-supervised approach.
\end{abstract}

\section{Introduction}
The widespread adoption of deep learning methods in computer vision owes its success to learning powerful visual representations \cite{bengio2013representation}; however, this was achievable only with intensive manual labeling effort (which is extravagant and not scalable). Therefore, unsupervised feature learning \cite{gidaris2018unsupervised,noroozi2017representation,noroozi2016unsupervised,lee2017unsupervised,zhang2017split,wang2015unsupervised,caron2018deep,li2018unsupervised,wu2018unsupervised} has recently been widely adopted to extract data representation without the need for such label information. This representation can be used for different tasks of image \cite{krizhevsky2012imagenet} or video classification \cite{karpathy2014large}.

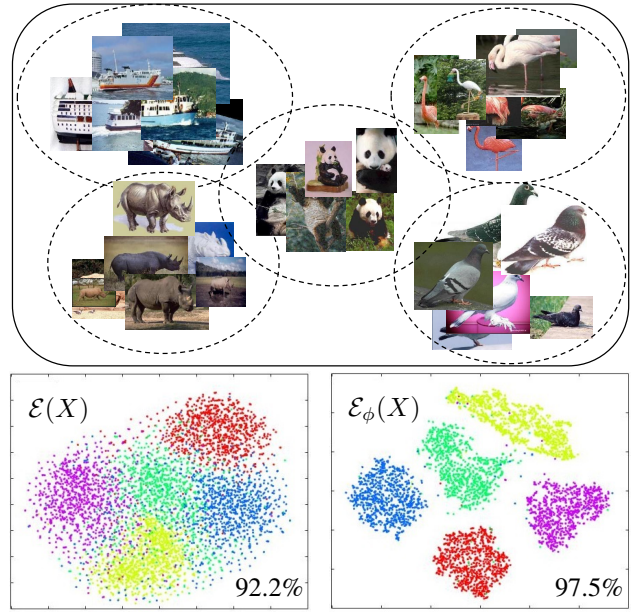
\begin{figure}[t]
\begin{center}
\includegraphics[width=\linewidth]{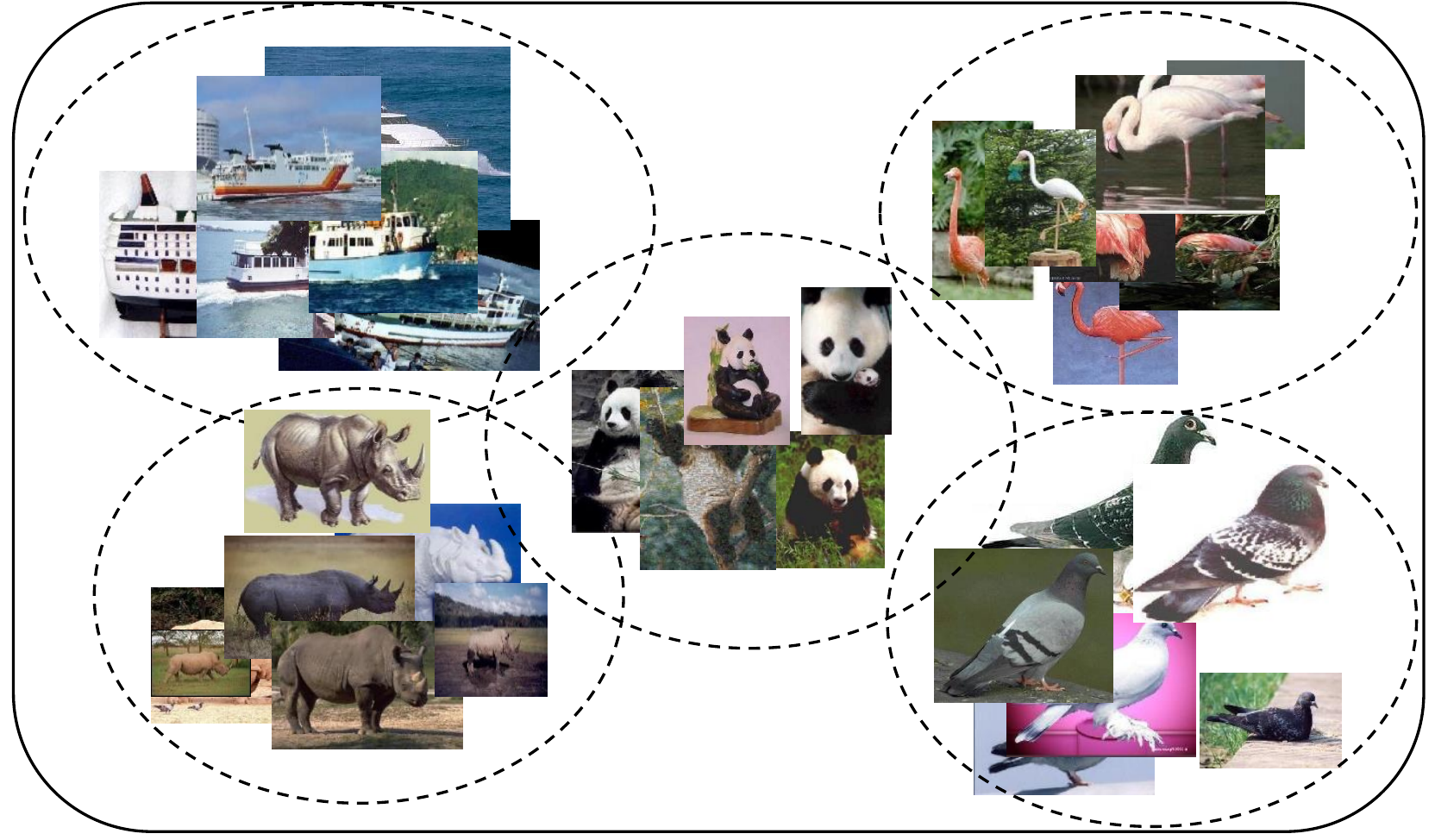}
\begin{tikzpicture}
    \node[anchor=south west,inner sep=0] at (0,0) {\includegraphics[width=0.48\linewidth]{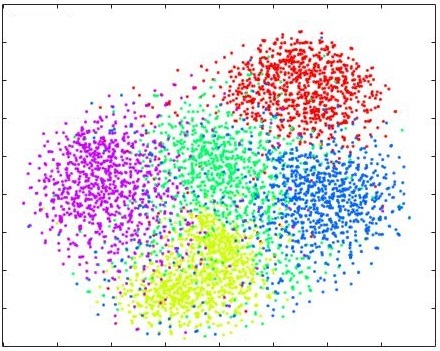}};
    \node[anchor=south west,inner sep=0] at (4.25,0) {\includegraphics[width=0.48\linewidth]{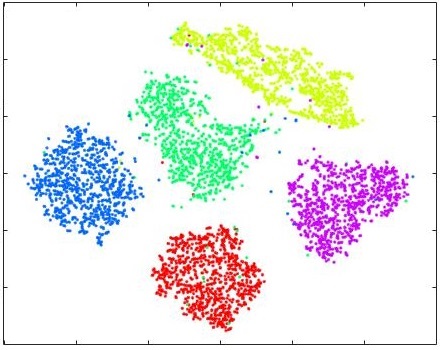}};
    \node[anchor=south west,inner sep=0] at (0.25,2.5) {$\mathcal{E}(X)$};
    \node[anchor=south west,inner sep=0] at (4.5,2.5) {$\mathcal{E}_{\phi}(X)$};
    \node[anchor=south west,inner sep=0] at (3,0.2) {92.2\%};
    \node[anchor=south west,inner sep=0] at (7.25,0.2) {97.5\%};
    
\end{tikzpicture}
\end{center}
   \caption{Some samples from five classes of the Caltech dataset (top), latent space visualization using a regular AE (\ie, $\mathcal{E}(X)$; left), and our proposed AE that encodes the neighborhood relations (\ie, $\mathcal{E}_\phi(X)$; right). With the same classifier, $\mathcal{E}(X)$ leads to a classification accuracy of 92.2\% and $\mathcal{E}_\phi(X)$ 97.5\%. }
 \label{fig:fig1}
\end{figure}

Unsupervised representation learning in the context of deep networks has often been defined by minimizing the reconstruction error \cite{makhzani2015adversarial}, such as in auto-encoders (AEs). 
AEs have shown to be great tools for unsupervised representation learning in a variety of tasks, including image inpainting \cite{pathak2016context}, feature ranking \cite{sharifipour2019unsupervised}, denosing \cite{vincent2008extracting}, clustering \cite{zhao2019variational}, defense against adversarial examples \cite{meng2017magnet}, and anomaly detection \cite{sabokrou2018avid,sabokrou2017fast}. Although AEs have led to far-reaching success for data representation, there are some caveats associated with using reconstruction errors as the sole metric for representation learning: (1) As also argued in \cite{wang2017feature}, it forces to reconstruct all parts of the input, even if they are irrelevant for any given task or are contaminated by noise; (2) It leads to a mechanism that entirely depends on single-point data abstraction, \ie, the AE learns to just reconstruct its input while neglecting other data points present in the dataset. The semantic relationship between neighboring samples in the dataset endure rich information that can direct learning more representative features (overlooked in current AE settings). 

To overcome the above challenges and enhance the performance of the popular encoder-decoder networks (\ie, AEs), in this paper, we propose a simple yet effective encoder-decoder architecture using self-supervised learning strategies. The self-supervised component encodes the neighborhood relations among the data points present in the training set. This setting goes beyond looking at the reconstruction of each data point separately, and self-supervises the model such that the conceived latent space preserves proper local neighborhood structures. Different from most previous works \cite{hinton2006reducing,makhzani2015adversarial}, which aim at preserving the global Euclidean structure, our proposed Neighborhood-Relational Encoding (NRE) aims at preserving the local neighborhood structure on the data manifold. As a result, we expect that NRE will be less sensitive to noise and outliers. Our proposed structure includes an encoder that also encodes neighborhood relations, denoted by $\EE_\phi$ (as opposed to $\EE$ in regular AEs), and a decoder, $\mathcal{D}$, which are jointly learned (similar to an auto-encoder). Therefore, $\EE$ encodes the input sample $X$ to a discriminative latent space $\mathcal{R}$, from which $\mathcal{D}$ must be able to retrieve the original sample. To learn the neighborhood relations, $\EE_\phi$ requires to operate as a kernel \cite{ham2004kernel} and map close-by data points closely to each other in the latent space (see Fig. \ref{fig:fig1}). 


In summary, the main contributions of this paper are as follows: (1) We propose a new learning strategy for $\EE$ncoder-$\mathcal{D}$ecoder deep network by introducing NRE. To the best of our knowledge, this article is the first to present an encoding network that simultaneously learns kernels (neighborhood cues) among inputs. (2) Leveraging the self-supervision injected as a result of our NRE component, we improve the performance of auto-encoders, which are popular tools of feature learning, (3) Our proposed scheme efficiently learns the semantic concepts within the visual data, and achieves state-of-the-art results on different applications, such as image classification, anomaly detection, and defense against attacks from adversarial examples.

\section{Related Work}
Unsupervised learning through learning representation space that successfully reconstructs samples is widely used for a variety of tasks, including classification \cite{krizhevsky2012imagenet}, de-nosing \cite{vincent2008extracting}, and in-painting \cite{pathak2016context}. 
Conventional methods for unsupervised representation learning are usually based on a pretext task such as reconstruction of static images \cite{noroozi2017representation} or videos \cite{wang2015unsupervised}. Learning to reconstruct data was used for tasks like de-nsoing \cite{vincent2008extracting}, in-painting \cite{pathakCVPR16context}, image refinement for defense against adversarial example \cite{samangouei2018defense}, and for one-class classifiers \cite{sabokrou2018adversarially,sabokrou2016video,sabokrou2018avid}. This paper focuses on a new way for training encoder-decoder networks incorporating self-supervised neighborhood constraints. In the following, we briefly survey recent un/self-supervised representation learning and learn-to-reconstruct methods.  
 
\noindent\textbf{Un/Self-Supervised Representation Learning:}
Learning with respect to a pretext task is the central idea for unsupervised representation learning. As mentioned, learning-to-reconstruct images is a common pretext task for unsupervised feature learning \cite{hinton2006reducing}. Earlier works were based on precisely reconstructing the input images. But recent work tried constructing other modes of the data alongside reconstructing images themselves. Some examples include constructing an image channel from another one \cite{zhang2017split}, colorizing gray-scale images \cite{zhang2016colorful,larsson2016learning}, and in-painting \cite{pathak2016context}. 

Other types of pretext tasks proposed for unsupervised learning include understanding the correct order of video frames \cite{brattoli2017lstm,misra2016shuffle} or predicting the spatial relation between image patches \cite{doersch2015unsupervised}, \eg, jigsaw puzzle solving as a pretext task was exploited by Noorozi and Favaro \cite{noroozi2016unsupervised}. In another work, Noroozi \etal~\cite{noroozi2017representation} proposed to train an unsupervised model by counting the primitive elements of images. Pathak \etal~\cite{pathak2017learning}  proposed a model to segment an image into foreground and background. Some methods use external signals that may come freely with visual data. For instance, some methods use known motion cues like ego-motion \cite{jayaraman2015learning,agrawal2015learning} or sound \cite{owens2016ambient} as sources for self-supervision. 
Most of these works ignored the relationship between samples. Some recent work \cite{agrawal2015learning,pathak2017learning} tried to model the relation between video patches as a pretext task. Conceptually, these works are related to our work, but different from our method, these pre-trained networks were developed for ad-hoc purposes, and were not capable of being applied to other computer vision tasks. Additionally, we introduce more comprehensive neighborhood cues to discover the intrinsic local manifolds. Che \etal~\cite{chu2017stacked} proposed a method for unsupervised feature learning using a similarity-aware auto-enocder that aims to map similar samples close to each other. However, unlike our method, they neglected the important relational information among the samples.

\noindent\textbf{Learning-to-Reconstruct:}
As discussed earlier, reconstruction can be considered as a pretext task for un/self-supervised representation learning. Many of computer vision tasks are dependent on this simple idea. There is a wide range of applications, but we briefly go over the tasks used for evaluation in this paper. 

Sabokrou \etal~\cite{sabokrou2018avid, sabokrou2018adversarially} used reconstruction errors and the reconstructed video frames for end-to-end one-class classification applied to anomaly detection. They analyzed the reconstruction error for detecting anomalies \cite{sabokrou2016video} the reconstructed (or refined) images to create better discrimination between normal and anomaly images \cite{sabokrou2018adversarially}. MagNet \cite{meng2017magnet} and Defense-GAN \cite{samangouei2018defense} as two important baseline for defense against adversarial attacks are based on refinement of adversarial examples using reconstruction techniques. MagNet directly refines the adversarial examples using an encoder-decoder trained on normal samples. Defense-GAN refines the adversarial example using a GAN generator that is trained only on normal images. The generator maps input examples to its latent space and generates images from the latent space that are hopefully free from the adversary.

\section{Method}
\begin{figure}[t]
\begin{center}
   \includegraphics[width=1\linewidth]{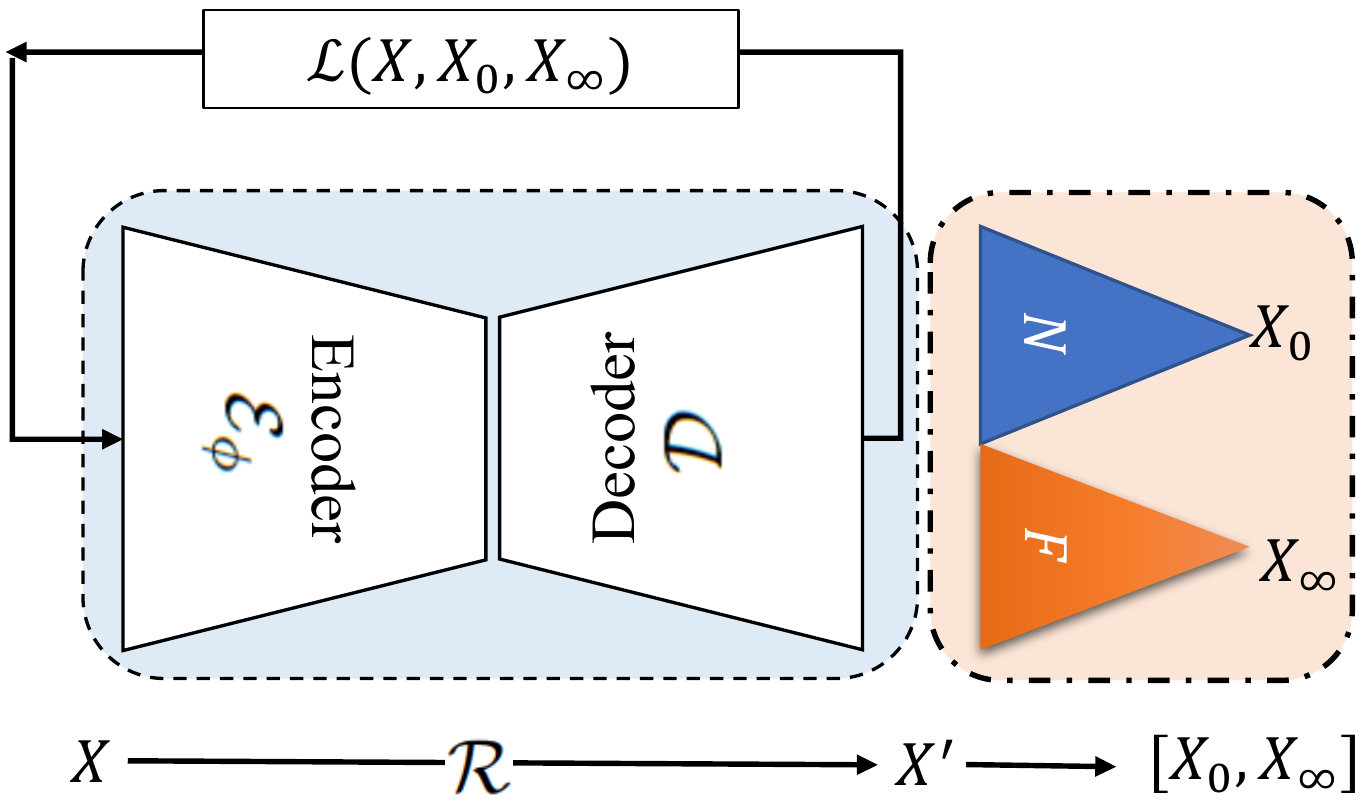}
\end{center}
   \caption{Overview of the proposed structure for self-supervised feature learning. $\mathcal{E}+\mathcal{D}$ are learned through a forward path and back-propagation of the error. In the forward path, $X'$ is retrieved from the input $X$, but a relational loss $(\mathcal{L}(X,X_0,X_\infty))$ is back-propagated to train the $\mathcal{E}+\mathcal{D}$ networks. $N$ and $F$ are the modules that identify $X_0$ and $X_\infty$ from the dataset.}
\label{fig:os}
\end{figure}

Our proposed approach for self-supervised representation learning is composed of three important components: (1) The encoder network $\EE$; (2) the decoder network $\mathcal{D}$; and (3) the objective function that incorporates the neighborhood relational information. The joint network $\EE+\mathcal{D}$ is trained as an encoder-decoder network based on the proposed objective function. $\EE$ provides a reduced representation $\mathcal{R}$ of its input sample $X$, with maximum information preservation, which enables $\mathcal{D}$ to retrieve $X$ from $\mathcal{R}$. The output of $\mathcal{D}$ is denoted by $\hat{X}$. Our goal is to train this reconstruction so that $\tilde{X}$ is similar not only to $X$ but also to its neighbouring sample(s) $X_0$, while being dissimilar to its far-away sample(s), \ie, $X_{\infty}$. This infrastructure concludes a self-supervised (and hence unsupervised) representation ($\EE(X)$) that can be used for any image or video analysis tasks. Encoding the neighborhood relations into the representation makes the learned feature space more separable. Fig.~\ref{fig:os} shows a sketch of our approach. $\EE$ and $\mathcal{D}$ are trained to discover the relationship among the samples. 

First, consider a setting that ${\EE+\mathcal{D}}$ defines an auto-encoder (AE) that is pre-trained to only reconstruct the input sample, \ie, reconstruct $X$ and obtain $X'$. Using this pre-trained network, we propose a procedure to identify $X_0$ and $X_\infty$ based on the latent space of ${\EE+\mathcal{D}}$, $\RR$, using the two modules, $N$ and $F$, respectively. Then, we optimize the network parameters of $\EE$ and $\mathcal{D}$ jointly using a loss function $\mathcal{L}(X,X_0,X_\infty)$, in which the neighborhood-relational information is propagated with respect to sample $X$. We denote our AE encoder that incorporates neighborhood information for building the latent space as $\EE_\phi$. As the training continues, ${\EE_\phi+\mathcal{D}}$ learns to better encode neighborhood information into $\RR$ and hence $N$ and $F$ can better uncover close-by and far-away samples. After training the network, $\mathcal{R}$ (\ie, ${\EE}_\phi(X)$) provides a discriminative representation of $X$. Furthermore, $\mathcal{D}\left(\EE_\phi(X)\right)$ acts as a refiner for $X$ regularizing the reconstructed $X'$ by its neighbors, which can be integrated as a pre-processing step for different classification or regression tasks. $X'$ is the reconstructed version of $X$ in the training phase, while  ${\tilde{X}}$ is its  reconstructed version using the trained relational AE. Detailed descriptions of each module and the overall training/testing procedures are described in the following subsections. 

\begin{figure}[t]
        \centering
 \includegraphics[width=1\linewidth]{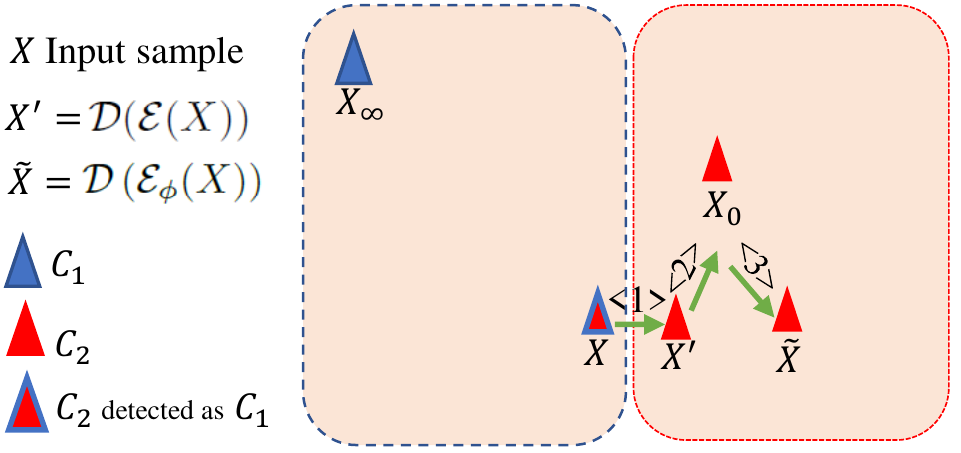}    
        \caption{A schematic sketch of procedures for reconstructing $X$ in 2D space. Suppose we have two classes of data (\textcolor{blue}{blue} $C_1$ and \textcolor{red}{red} $C_2$). Let $X \in C_2$, but being mistakenly classified as $C_1$. Here, we analyze and apply terms of the loss function in \eqref{eq:l1} one by one. Based on term Eq.~(1), $X$ is supposedly transferred to (reconstructed as) $X'$ using $<$1$>$; the second and third terms ($<$2$>$ and $<$3$>$) bring $X'$ closer to $X_0$ and farther from $X_\infty$, respectively. As can be seen, the final reconstructed sample ($\tilde{X}=\mathcal{D}(\EE_{\phi}(X)$)) may be put on the correct side.} 
        \label{fig:schematic}
\end{figure}

\subsection{Neighborhood-Relational Encoding (NRE)}
Traditional unsupervised representation learning highly depends on a pretext assumption, often defined on top of the reconstruction power of the learned features. These methods learn the spatial dependencies within the images and hence the inter-relations among the data points are neglected. As shown by methods that operate in the neighborhood spaces, such as $K$-nearest neighbour \cite{cover1967nearest}, as a rule of thumb, samples that are spanned close-by in the space of all samples tend to belong to the same classes. Also, kernel methods \cite{shawe2004kernel} suggest that modifying the representation space based on the (positive-definite) similarities generally leads to more dicriminative and separable spaces. Encoder-decoder networks have also been investigated for such properties, \eg, in \cite{sabokrou2018adversarially} where it was shown that samples can be efficiently refined and be made more separable for anomaly classification tasks. Inspired by the previous work, we propose an encoder-decoder AE deep network to learn representations of the data through self-supervision derived from the neighborhood cues in the data manifold. 

To this end, we force $\mathcal{D}\left(\EE_\phi(X)\right)$ to reconstruct $X$ mindful of its neighbor(s) $X_0$, while trying to distant from the far-away sample(s) $X_\infty$. Therefore, the parameters of $\EE_\phi+\mathcal{D}$ are learned using the following loss function:
\begin{align}
    \mathcal{L}= & \underbrace{\lambda_1\mathbb{D}( \mathcal{R}_\mathcal{A}(X),\mathcal{R}_\mathcal{A}(X')) }_{<1>} 
     +\underbrace{\lambda_2\mathbb{D}(\mathcal{R}_\mathcal{A}(X'),\mathcal{R}_\mathcal{A}(X_0))}_{<2>} \nonumber \\
    & +\underbrace{\lambda_3\mathbb{S}(\mathcal{R}_\mathcal{A}(X'),\mathcal{R}_\mathcal{A}(X_\infty))}_{<3>}, \label{eq:l1}
\end{align}
where $X'=\mathcal{D}({\EE}(X))$, $\lambda_{i\in\{1,2,3\}}$ are scaled regularization hyperparameters with $\lambda_1+\lambda_2+\lambda_3=1$, $X_0$ and $X_\infty$ are calculated by $N(X)$ and $F(X)$, respectively. $N(\cdot)$ and $F(\cdot)$ define functions that return the closest and the farthest samples to their inputs, respectively (defined in detail later). $\mathbb{D}$ and $\mathbb{S}$ are two metrics for computing the distance and similarity of two vectors, and $\mathcal{E}_\phi(X)$ is the new representation of $X$. To obtain the similarities and distances of samples we use a pre-trained AE network. We denote the encoder of this AE as $\mathcal{A}$ and its latent space as $\mathcal{R}_\mathcal{A}$. $\mathbb{S}$, $\mathbb{D}$,  and $\mathcal{R}_\mathcal{A}$ are  explained in more details in the following subsections. After training this network, $\mathcal{R}=\mathcal{E}_\phi(X)$ will be a representation of $X$, which is forced be similar to the representation of the closest sample(s) and dissimilar to far-away one(s). 
Fig.~\ref{fig:schematic} shows how sample $X$ is refined in a 2D space with respect to each term in the loss function, Eq.~\eqref{eq:l1}. As can be seen, after refining, $X$ moves closer to the center of its correct class. Note that $X_0$ and $X_\infty$ should be calculated with respect to $X'$, not $X$. 
\vspace{-15 pt}
\paragraph{Neighboring Relation}
\label{sec:nr}
As shown in Fig.~\ref{fig:os}, there are two important modules, $D$ and $F$, with key roles that provide side information for joint training of $\EE_\phi+\mathcal{D}$. As mentioned earlier, $D$ and $F$ are defined to find the closet or most similar sample(s) and far-away or most dissimilar sample(s), respectively. There are several measures to infer the similarity of two samples (\eg, images) in an unsupervised fashion. Direct image similarity methods, such as SSIM \cite{wang2004image}, are too high-level and often fail to evaluate the semantics of the image. Hence, instead of directly working with images in the original space, we compare them in the latent representation space. To this end, an encoder network, $\mathcal{A}$, is trained on all unlabeled available samples to provide a discriminative representing of the samples. The encoder unsupervisedly and jointly with a decoder, is trained to form an auto-encoder. Let $\mathcal{X}=\{{X_i}\}_{i=1}^{i=Z}$ be our dataset with size $Z$ and $\mathcal{R}_\mathcal{A}(X_i)$ be the corresponding representation on $X_i$ using $\mathcal{A}$. 

Under the above setting, $X_0$, the closest sample to $X'$ is calculated using
\begin{equation}
X_0=N(X')=\argmax_{X_i\in \mathcal{X}, X_i \neq X'} \mathbb{S}(\mathcal{R}_\mathcal{A}(X_i),\mathcal{R}_\mathcal{A}(X')),
\label{eq:cl}
\end{equation}
and $X_\infty$, the most dissimilar sample to $X$ is defined as:
\begin{equation}
X_\infty=F(X')=\argmin_{X_i \in \mathcal{X}, X_i \neq X'} \mathbb{S}(\mathcal{R}_\mathcal{A}(X_i),\mathcal{R}_\mathcal{A}(X')).
\label{eq:far}
\end{equation}
In addition, $\mathbb{D}(\cdot,\cdot)=1-\mathbb{S}(\cdot,\cdot)$ and $\mathbb{S}(\cdot,\cdot)$ is a cosine similarity measure computed by:
\begin{equation}
\label{eq:d}
   \mathbb{S}(\mathbf{a},\mathbf{b})= {\mathbf{a} \cdot \mathbf{b} \over \|\mathbf{a} \|\|\mathbf{b} \|}={\frac {\sum_{i=1}^{n}{a_{i}b_{i}}}{{\sqrt {\sum_{i=1}^{n}{a_{i}^{2}}}}{\sqrt {\sum_{i=1}^{n}{b_{i}^{2}}}}}},
\end{equation}
where $a_i$ and $b_i$ are the $i^\text{th}$ components of vectors $\mathbf{a}$ and $\mathbf{b}$, respectively. Without loss of generality, we can select a set of (more than one) similar and dissimilar samples. $\mathcal{A}$ is composed of several convolutional, sub-sampling, and fully connected layers 
There is a ReLU layer on top of $\mathcal{R}_\mathcal{A}$, which forces all of its output values to be positive. Consequently, always $\mathbb{S}(\mathcal{R}_\mathcal{A}(X_i),\mathcal{R}_\mathcal{A}(X'))$ is positive. 

\subsection{Training $\EE_\phi+\mathcal{D}$}
\label{sec:edlearning}
These two networks are jointly trained. Training sample $X$ is fed to $\EE_\phi$+$\mathcal{D}$, which creates an output $X'$. The network is optimized using the  loss function $\mathcal{L}$ (Eq.~\eqref{eq:l1}). This turns $X$ to a more discriminative sample based on the neighborhood encoding scheme (see Subsection \ref{sec:nr}). Eq.~\eqref{eq:l1} considers only one nearby and one far-away sample, but for robustness against outliers and to better discover the relationship of samples, the network can be trained using a set of such samples (more than one). Therefore, the loss function could be re-written as follow: 

\begin{equation}
\begin{aligned}
        \mathcal{L}=&\, \lambda_1\mathbb{D}(\mathcal{R}_\mathcal{A}(X),\mathcal{R}_\mathcal{A}(X')) \\ & +  \lambda_2 \sum_{i=1}^T{\mathbb{D}(\mathcal{R}_\mathcal{A}(X'),\mathcal{R}_\mathcal{A}({X_0}_i))} \\ & +\lambda_3\sum_{i=1}^T{\mathbb{S}(\mathcal{R}_\mathcal{A}(X'),\mathcal{R}_\mathcal{A}({X_\infty}_i))},
\end{aligned}
\end{equation}
where  $T$ is a hyperparameter denoting the number of  selected nearby/faraway samples. Generally, greater $T$ concludes better performance, but its side-effect is an expensive training phase, and if a very large $T$ is selected, then the set of far-away and nearby samples may have common elements, which is not desirable. Note that finding $X_0$ and $X_\infty$ is a time consuming task, which is proportional to the the size of the training set. To cope with this, we cluster the training samples into $K$ clusters and the nearest neighbour sample(s) are selected from the same cluster of $X$, and faraway samples are randomly selected from clusters that have faraway centers from the cluster to which $X$ belongs. This simple technique speeds up the training process drastically. Furthermore, instead of training $\EE_\phi+\mathcal{D}$ from scratch, the wights of these networks are initialized based on an optimized traditional encoder-decoder network. 

\begin{figure}
    \centering
 \includegraphics[width=\linewidth]{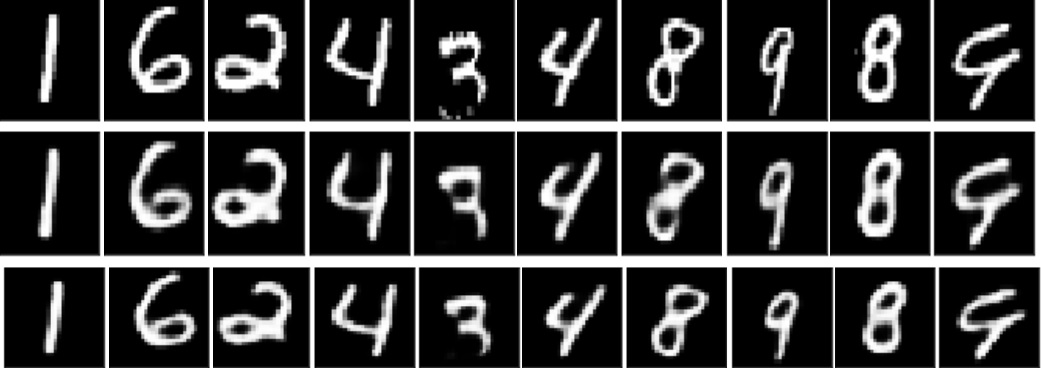}    
 \caption{Several examples of reconstructed images from the original MNIST sample (1$^{st}$ row) using a conventional AE (3$^{rd}$ row) and our proposed encoder-decoder network (2$^{nd}$ row) . Conventional network is optimized based on the reconstruction errors and our network is trained to optimize the  reconstruction error alongside of neighborhood-relational information, Eq.~\eqref{eq:l1}. }
    \label{fig:rec}
\end{figure}

The hyperparameters $\lambda_i$ have a key role on the final performance of the network and can be set dependent on application. After a joint training of the ${\EE}_\phi$+$\mathcal{D}$, and with respect to the values of $\lambda_{i\in\{1,2,3\}}$, the networks can be interpreted as the following:
\begin{itemize}
    \item $||X-\mathcal{D}(\EE_\phi(X))||^2 <\epsilon_1, ||X_0-\mathcal{D}(\EE_\phi(X))||^2 <\epsilon_2$, where $\epsilon_{1}$ and $\epsilon_{2}$ are small non-negative scalars. But $||X_\infty-\mathcal{D}(\EE_\phi(X))||^2 >\epsilon_3$, where $\epsilon_3$ is very larger than  $\epsilon_{1}$ and $\epsilon_{2}$. As aforementioned, $X_0$ and $X_\infty$ are close to and far from  $X$, respectively. Consequently, we can say that with a high probability $X_0$ and $X$ come from the same class and $X_\infty$ from another class. Accordingly, $\mathcal{D}({\EE_\phi}(X))$ is forced to be close to the samples from the same class and away from samples of other classes, leading to more separable samples in the reconstructed space.
 
    \item  Let  $\mathcal{P}_{c}$ be the probability of  an specific classifier labeling $X$ as class $c$. We expect that  $\mathcal{P}_{c}({\EE_\phi}(X))|_{\frac{1}{3},\frac{1}{3},\frac{1}{3}} > \mathcal{P}_{c}(\EE_\phi(X))|_{1,0,0}$. The subscripts $|_{\lambda_1,\lambda_2,\lambda_3}$ denote the values of $\lambda_1, \lambda_2,$ and $\lambda_3$, respectively. This is because ${\EE_\phi}$ is forced to map  sample $X$ to a latent-space with enough neighborhood information that will result in a more separable decoding. 
    
    \item In a classification problem, if $X$ belongs to the class $c$, and $\lambda_2$ and $\lambda_3$ are selected large enough, it is expected that $\mathcal{P}_{c}({D}(\EE_\phi(X+\sigma)))|_{\lambda_1,\lambda_2,\lambda_3}>\mathcal{P}_{c}({D}(\EE_\phi(X+\sigma)))|_{1,0,0}$, where $\sigma$ denotes noise element. Our model considers the relation of sample $X$ with its neighbors to make the model robust against noise and outliers. Similar concept is investigated in \cite{he2005neighborhood}. This characteristic of our relational reconstruction is an effective mechanism for defense against adversarial attacks. To defend deep networks against adversarial example attacks, reconstruction of adversarial examples using our encoder-decoder formulation trained on original (clean and normal) sample set is very useful. Similar argument can be found in recent defence mechanisms such as MagNet \cite{meng2017magnet} and defense-GAN \cite{samangouei2018defense}.
    
    \item $||X-\mathcal{D}(\EE_\phi(X))|_{\lambda_1,\lambda_2\neq0,\lambda_3\neq0}||^2 \approx  ||X-\mathcal{D}(\EE_\phi(X))|_{1,0,0}||^2$. This implies that although our formulation (\ie, learning to reconstruct an example with respect to neighborhood and relational information) does not just focus on the reconstruction, after training it is still able to efficiently reconstruct the input samples (see Fig. \ref{fig:rec}). Furthermore, our formulation does not over-emphasize on reconstruction loss only and borrows information from the neighborhood embedding. Therefore, it reconstructs data based on semantics instead of pixel value loss functions. 
\end{itemize}

We know that the relational information contains important cues, but paying extra attention to them and not properly preserving the context of samples (\ie, the reconstruction error) may conclude adverse results. Normally, the relational information is exploited besides context, as a side-information.~Accordingly, to create a trade-off between these two sources of information, we set $\lambda_1>\lambda_{2},\lambda_{3}$.

\section{Experimental Results}
In this section,  the proposed method is evaluated on different datasets and tasks, to showcase its reliability and generality.  The performance results are analyzed in details and are compared with state-of-the-art techniques. To show the adaptability and generality of the introduced framework for a wide range of applications it is evaluated as (1) an auto-encoder (R-AE), (2) a self-supervised (unsupervised) representation learning method, (3) as a defense approach against adversarial example attacks, and finally for (4) anomaly detection. Our results are at least comparable or better than state-of-the-art methods in each of these fields.

\subsection{Setup}
Several deep networks are exploited in our experiments, which are explained in details in the supplementary material\footnote{More details at: \url{https://github.com/Sabokrou/NRE}}.  
The weights of network $\EE_\phi+\DD$ are initialized based on the Adam optimizer and learning rate is set to 0.0001. 
Depending on the task, $\lambda_1,\lambda_2,$ and $\lambda_3$ are selected and are shown by a triplet (${\lambda_1,\lambda_2,\lambda_3}$), subscripted for each method. All the reported results in this section are from our implementation using Tensor-Flow framework \cite{abadi2016tensorflow} and Python ran on an NVIDIA TITAN X.

\subsection{Unsupervised Learning with NRE}
Conventional formulations of AE were widely used as a popular tool for  unsupervised feature learning. In the recent years, new versions of AE such as split-brain \cite{zhang2017split}, adversarial auto-encoder \cite{makhzani2015adversarial}, and context auto-encoder \cite{pathakCVPR16context} were presented. We evaluate the performance of our method (NRE) on MNIST dataset \cite{lecun1998mnist} and compare the results with these various types of AE. 
{MNIST\footnote{Available at http://yann.lecun.com/exdb/mnist/}} dataset  includes 60,000 handwritten digits from `0' to `9', with 50,000 and 10,000 samples  as training and testing data, respectively. 

\paragraph{Results on MNIST}
To evaluate the performance of various version of AEs, auto-encoders are trained with respect to different policies  (objective functions). After training these networks, all training and testing samples are mapped to the AE latent representation space based on trained auto-encoders. Equally for each AE, a Support Vector Machine (SVM) \cite{vapnik1995support} classifier on top of the represented training samples is trained and the classification accuracy on test set is reported. The results are shown in Table \ref{tab:ae}. As can be seen, the classification accuracy based on  relational encoding is better than other methods. For fair comparison, all AEs are trained for 40 and 100 epochs. The classification is done by both a Linear SVM (L-SVM) and one with an RBF kernel (R-SVM). The hyperparameter of RBF is set as 0.01 and is fixed in all experiments. 

Our proposed objective function is a more complex one compared to the conventional AE, and therefore it will achieve better results when it is trained for more epochs (even more than 100 epochs). But for fairness, all methods are trained for the same number of epochs. 
\begin{table}[t]
    \caption{Compassion results of the accuracy for our method (NRE) with conventional and widely-used auto-encoders. The best results are typeset in bold. NRE is subscripted with the chosen hyperparameters (\ie, NRE$_{\lambda_1,\lambda_2,\lambda_3}$).}
    \centering
    \begin{tabular}{lcccc}\hline
Classifer & \multicolumn{2}{c}{L-SVM} &  \multicolumn{2}{c}{R-SVM} \\ 
\# of Epochs   & 40&100 &   40&100  \\\hline \hline
 AE \cite{hinton2006reducing} &  0.969&0.969 &   0.961&0.972  \\
 DAE \cite{vincent2008extracting}  &  0.942&0.936  & 0.954&0.964 \\
 Context AE \cite{pathakCVPR16context} &0.970&0.974 & 0.978&0.981 \\
 Split-Brain AE \cite{zhang2017split} & 0.972&0.973&  0.975&0.979\\ 
 NRE$_{0.5,0.2,0.3}$ (Ours)& \textbf{0.977}&\textbf{0.978} &  \textbf{0.981}&\textbf{0.984} \\ \hline
    \end{tabular}
    
    \label{tab:ae}
\end{table}
\begin{table}[t]
    \centering
    \caption{Performance of our self-supervised representation based on NRE for classification, detection, and segmentation tasks. Classification and Fast R-CNN \cite{girshick2015fast} detection results for the PASCAL VOC 2007 \cite{everingham2015pascal} test set, and FCN \cite{long2015fully} segmentation results on the PASCAL VOC 2012 validation set, under the standard mean average precision (mAP) or mean intersection over union (mIOU) metrics for each task. Classification, Det and Seg columns show classification, detection, and segmentation results, respectively.  }
    \label{tab:un}
    \setlength{\tabcolsep}{3.5pt}
    \begin{tabular}{lcccccc} \hline 
    & \multicolumn{3}{c}{Classification} & ~& Det& Seg  \\
 \cline{2-4} \cline{6-6} \cline{7-7}
      Layers &  FC8 & FC6-8 &all & & all  & all\\ \hline
      AlexNet \cite{krizhevsky2012imagenet} & 77.0& 78.8 &78.3 & & 56.8  &48.0\\ 
    Agrawal \etal~\cite{agrawal2015learning}& 31.2& 31.0 &54.2 & & 43.9  &-- \\
Pathak \etal~\cite{pathakCVPR16context}&30.5 &34.6 &56.5 & &  44.5  &30.0\\
Wang \etal \cite{wang2015unsupervised}~  & 28.4 &55.6 &63.1 & & 47.4  &-- \\
Doersch \etal~\cite{doersch2015unsupervised}&{44.7}& 55.1& 65.3 & & 51.1  & -- \\
K-means \cite{krahenbuhl2015data}& 32.0 &39.2 &56.6 & &  45.6 & 32.6 \\
AE \cite{hinton2006reducing}& 24.8 &16.0 &53.8 & & 41.9 & --\\
BiGAN \cite{donahue2016adversarial}& 41.7 &52.5 &60.3& & 46.9 &  35.2\\
Counting \cite{noroozi2017representation}&--&--&67.7 &  &51.4  &36.6\\
Owens \etal~\cite{owens2016ambient} &--&--&61.3 & &44.0 & -- \\
Pathak \etal\cite{pathakCVPR16context}&--&--& 61.0 & &52.2 &-- \\
Jenni \etal \cite{jenni2018self}&--&--&69.8 & &52.5& 38.1\\
 DeepCluster \cite{caron2018deep} &--&--&73.7  & &55.4  & 45.1\\ 
 Noorozi \& Favaro \cite{noroozi2016unsupervised} &--&--&67.6&&53.2&37.6\\ 

NRE$_{0.5,0.25,0.25}$ (Ours) &\textbf{55.9}&\textbf{71.2}&\textbf{74.4}& & \textbf{54.7} & \textbf{51.1} \\

         \hline 
    \end{tabular}
    
\end{table}
\subsection{Classification, Detection, and Segmentation}
As mentioned before, unsupervised or self-supervised representation is increasingly used for different applications because of its advantage of not requiring labeled data. A pretext task is often first trained to direct the ultimate network to have proper initialization or even create the embedding space for the subsequent task. We compare our approach to the state-of-the-art methods, which all use variants of AlexNet \cite{krizhevsky2012imagenet}. We follow \cite{zhang2017split}, for evaluating and comparing our method with the others.     
We pre-trained our network to learn the relational information on the ImageNet dataset \cite{deng2009imagenet}. This dataset is very large, so finding $X_0$ and $X_\infty$ is very expensive and time-consuming. To end this, as mentioned in Section \ref{sec:edlearning}, the  dataset is divided  to $K=400$ clusters and then just the partition involving any specific $X$ is searched for finding its $X_0$ and $X_\infty$ is randomly selected from clusters with faraway centers
from the cluster of $X$. We evaluate the performance of our relational representation on PASCAL VOC dataset \cite{everingham2010pascal} as a  benchmark set for classification tasks. This classiﬁcation task involves 20 binary classiﬁcation decisions regarding the presence or absence of 20 object classes. We used the AlexNet architecture and embeded it as the decoder in our AE formulation. We mirrored same architecture for the decoder by converting the convolutional and sub-sampling layers to de-convolutional and up-sampling layers. 

\paragraph{Results on PASCAL VOC}
Several classifiers  are trained by freezing various parts of the AlexNet \cite{krizhevsky2012imagenet}.  In the first experiment, on top of FC6 and FC7, a linear classifier is trained. In the second experiment, all three FC6, FC7 and FC8 layers are trained in a supervised manner, where all other layers were frozen. Finally, the entire network is `fine-tuned'. Table \ref{tab:un} compares our results in comparison with the state-of-the-art methods. 

We further evaluated object detection and segmentation tasks with the pre-trained AlexNet used as the initialization for Fast R-CNN \cite{girshick2015fast} and fully convolutional network (FCN) \cite{long2015fully}, object detection and segmentation tasks, respectively. For these tests, we replaced the supervised trained AlexNet \cite{krizhevsky2012imagenet} with our self-supervised trained network, as a pre-training for the specific task. Results confirm that the proposed method can be used as an efficient approach for self-supervised  feature learning. In all cases (except for the segmentation task) our results are superior to others by a considerable margin.   

\subsection{Defense Against Adversarial Attacks}
Adversarial examples are means of fooling trained networks for specific computer vision and are a challenging problem with respect to the safety of deep networks. Let $\mathcal{F}$ be a classifier, which has correctly labeled $X$ as $Y$, \ie, $\mathcal{F}(X)=Y$. Adversarial attack is done by contaminating $X$ in a way that leads to creating its equivalent adversarial example $\hat{X}$, where $||X-\hat{X}||<\epsilon$ ($\epsilon$ is a small non-negative scalar) and $\mathcal{F}(\hat{X})\neq Y$. This defines a vulnerability for the classifier $\mathcal{F}$  \cite{samangouei2018defense,meng2017magnet}.

As a defense mechanism, MagNet \cite{meng2017magnet} has proposed to refine the adversarial example using an auto-encoder using the manifold distribution of the correct class. Here, we show that our proposed encoder-decoder with NRE performs better than MagNet. We evaluate  our method and MagNet \cite{meng2017magnet} with respect to Fast Gradient Sign Attack (FGSM) \cite{papernot2016cleverhans} attack with different values of $\epsilon$. These experiments are done on the MNIST dataset.  We select 50,000 examples for the training set, 150 samples for training the substitute network, and finally 9850 samples for testing. In a black-box attack, the attacker does not have access to the architecture and weights of target classifier. But it is possible to emulate the behaviour of target classifier using a substitute network, learned on 150 samples. We trained a CNN network as our target classifier and obtained an accuracy of 98.6\%, alongside this classifier, a substitute CNN  classifier is trained on 150 samples with a 77\% accuracy. 
\begin{figure}
    \centering
    \includegraphics[width=\linewidth]{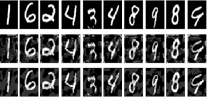}
    \caption{Some samples of adversarial examples created by FGSM \cite{papernot2016cleverhans} attack ($\epsilon=0.2$). First row: Original images; Second row: reconstruction of the adversarial examples by our method; Last row: the adversarial example.}
    \label{fig:attack}
\end{figure}

\vspace{-6pt}
\paragraph{Results of Defense Against Attacks}
We evaluate the performance of NRE as a defense approach against the adversarial examples and report the results in  Table \ref{tab:defense}. This table shows our results in comparison to MagNet \cite{meng2017magnet} as a baseline, which is based on reconstructing the adversarial sample. The architectures used for the auto-encoders of both methods are the same, 
but are learned with different object functions. A FGSM attack with different amounts of $\epsilon$ is used. Several adversarial examples and their reconstructions  by NRE are shown in Fig. \ref{fig:attack}. Naturally, after the attacks the accuracy of network is decreased. Our method and MagNet are applied on adversarial examples to refine them as a pre-processing step for the classifier. It can be seen that the accuracy of our method for all values  of $\epsilon$ is better than MagNet by a wide margin.   
\begin{table}[t]
\caption{Performance evaluation of NRE for refinement of adversarial examples as a defense strategy against adversarial attacks made by the black-box FGSM \cite{papernot2016cleverhans} attack. Best results in each column are typeset in bold.}
    \setlength{\tabcolsep}{4pt}
\begin{tabular}{llllll} \hline
$\epsilon$ & 0.01 &0.1 &0.2 &0.3 \\\hline \hline
MagNet \cite{meng2017magnet}&0.9238&0.7655& 0.614&0.4242\\
NRE$_{0.6,0.2,0.2}$ (Ours) &\textbf{0.9802}&\textbf{0.9056}&\textbf{0.8489}&\textbf{0.7166}\\\hline
\end{tabular}
\label{tab:defense}
\end{table}

There are more types of attacks and defense strategies against adversarial attacks, such DefenseGAN \cite{samangouei2018defense}. Analyzing all of these attacks and defense methods requires a deep discussion, which is out of the scope of this paper. Here, we briefly compared our method to the state-of-the-art to showcase efficiency of our method. 

\subsection{Video Anomaly Detection}
Detecting Anomalous events in videos (also refered to as irregularity detection in visual data) is an important task in different computer vision application. Recent state-of-the-art methods for anomaly detection are often based on encoder-decoder networks and analyzing the reconstruction error. As the context of our network  is very close  to these solutions for anomaly detection, we evaluate our method on this task as well.  

We evaluate our method on the UCSD Ped2 dataset \cite{chan2008ucsd}, which is a popular dataset for this task.  We follow the evaluation criteria of \cite{sabokrou2018adversarially}. Similar to \cite{sabokrou2018adversarially}, the frame-level accuracy is reported as the performance metric. In frame-level measure, a frame is considered as anomaly, if at least one of its pixels is detected as anomaly. The UCSD dataset has two subsets, referred to as Ped1 and Ped2. They are from different static-camera outdoor scenes, with 10 fps and resolutions of 158$\times$234 and 240$\times$360, respectively. Moving objects in these videos are mainly pedestrians, and all other objects like cars, wheelchairs, and bicycles are labeled as anomaly. To compare with the previous work on this dataset, we evaluate our algorithm on Ped2.

\vspace{-6pt}
\paragraph{Anomaly Detection Results}
For this experiment, we divided the  video frames into 2D patches of size 30$\times$30. All patches extracted from normal frames are considered for our training. Note that training data only contains normal patches. An encoder-decoder network with our objective function on all training patches is trained (See subsection \ref{sec:edlearning}). When the training is completed, test patches are fed to this encoder-decode network one by one. Similar to \cite{sabokrou2016video}, regular reconstruction error $(||X-\hat{X}||^2)$ is used as a measure for detecting the anomalies. If this reconstruction error is larger than a threshold, it means that the patch contains something that was not seen during training (\ie, it is an anomaly). Our method is very similar to \cite{sabokrou2016video}, but with two major differences: (1) We use only one auto-encoder but \cite{sabokrou2016video} has exploited two auto-encoder; (2) Our auto-encoder is learned based on relational-information, while \cite{sabokrou2016video} only trained based on reconstruction error. 

\begin{table}[t]
\caption{Frame-level anomaly detection comparisons on UCSD Ped2 in terms of Equal Error Rates (EER).} 
\begin{tabular}{ll|ll}
\hline
Method&EER&Method &EER    \\
\hline\hline
IBC~\cite{boiman2007detecting} & 13\%    & RE  \cite{sabokrou2016video} & 15\%  \\
MPCCA  ~\cite{kim2009observe} & 30\%   &  {\scriptsize Ravanbakhsh \etal}~\cite{ravanbakhsh2017training} &13\%\\
MDT ~\cite{mahadevan2010anomaly} & 24\% & {\scriptsize Ravanbakhsh \etal}~\cite{ravanbakhsh2017abnormal}& 14\%  \\
 {\scriptsize  Bertini \etal} ~\cite{bertini2012multi}& 30\% & {\scriptsize Dan Xu~\etal}~\cite{xu2015learning}& 17\%  \\
  Deep-Anomaly\cite{sabokrou2018deep} &13.5\% &  {\scriptsize Sabokrou \etal}~\cite{sabokrou2015real} &19\% \\
Li \etal~\cite{li2014anomaly}&18.5\% &    Deep-cascade\cite{sabokrou2017deep} &  \textbf{9\%}   \\
AVID \cite{sabokrou2018avid} &14\%&ALOCC\cite{sabokrou2018adversarially} & 13\%\\
\hline
 NRE$_{0.6,0.2,0.2}$& 17.5\%  & NRE$_{0.6,0.4,0}$ & 14\%  \\
\hline
\end{tabular}
\label{tab:EER}
\end{table}

Table \ref{tab:EER} reports the results of our method and other baseline and state-of-the-art approaches. Last row shows the results of our method with two different values for $\lambda_{1}$, $\lambda_{2}$, and $\lambda_{3}$. As can be seen, our method is comparable with the state-of-the-art. NRE is a very simple method based on the sole criteria of reconstruction error and neighborhood relational encoding, while other methods (such as \cite{sabokrou2015real}  and \cite{sabokrou2017deep}) are based on intensive spatio-temporal embedding of video content. The experiments show the generality of the proposed approach. We report our results with respect to different values of $\lambda_{1}$, $\lambda_{2}$, $\lambda_{3}$, and $T=1$.
 
\section{Discussion}
The results confirm that the proposed neighborhood relational encoding method for learning unsupervised and self-supervised representation can be adapted for a variety of computer vision and image analysis tasks. There are several challenges and interesting intuitions around NRE, which are discussed in the following:

\vspace{-9pt}
\paragraph{Values of $\lambda$:} The objective function consists of three terms, which can be adjusted depending on the target task. Our results show that $\lambda_1$ is very important on all types of tasks. But for tasks such as de-nosing, in-painting,  and generally enhancing the images, $\lambda_1$ and $\lambda_2$ show to be more important than $\lambda_3$. For classification and clustering tasks that require creating discriminative embedding spaces equal values of $\lambda_1$ and $\lambda_2+\lambda_3$ often result in better performance. 

\vspace{-9pt}
\paragraph{The hyperparameter $T$:} $T$ is a very important hyperparameter for capturing the intrinsic neighborhood-relational-information. Obviously, the larger it is selected to be, the more robustness is added for the method against outliers. However, if it is set to be very large, the concept of neighborhood will be lost. Therefore, a compromise should be made for each specific task. 

\vspace{-9pt}
\paragraph{Dynamic $\lambda$s:} Scheduling the values of $\lambda_{1}$, $\lambda_{2}$, and $\lambda_{3}$ during training can be very useful and lead to speed-ups in convergence. This can be a very interesting direction for the future work, as designing a good scheduling for gradually changing of these parameters (while interacting with each other) is not a straightforward task. 

\vspace{-9pt}
\paragraph{Metrics for finding similar or dissimilar images:} The main difficulty of the proposed method is finding the similar and dissimilar samples to any target image. We tested a wide range of metrics and found that simple cosine similarity in the latent space was enough for this purpose. Comparing images in their original space (pixel values) is not an appropriate option here, as it neglects the important context and semantics embedded in images. Better metrics can improve the results and be developed for specific applications.

\section{Conclusion}
In this paper, we proposed a learning framework for encoder-decoder networks (\eg, auto-encoders), and adopted it for a wide range of computer vision tasks. Our proposed method encodes the neighborhood-relations information into the AE and turns it to a kernel embedding framework. Therefore, besides learning a reconstruction scheme, our AE preserves the local geometric manifold. This leads to dicriminative neighborhood-guided self-supervised representation learning that can be used in a variety of applications, since it does not require label information for training. We evaluated our models in different related applications including self-supervise (unsupervised) representation learning for classification, detection, and segmentation, as well as defense against adversarial example attacks and anomaly detection in videos. The results suggest that our method is superior to, or at least comparable with, the state-of-the-art specific to each application while being much simpler.

\noindent\textbf{Acknowledgement}
M. Sabokrou was in part supported by a grant from IPM (No. CS1396-5-01). E. Adeli would like to thank Panasonic for the support.

{\small
\bibliographystyle{ieee_fullname}
\bibliography{refs}

\begin{thebibliography}{10}\itemsep=-1pt

\bibitem{abadi2016tensorflow}
Mart{\'\i}n Abadi, Paul Barham, Jianmin Chen, Zhifeng Chen, Andy Davis, Jeffrey
  Dean, Matthieu Devin, Sanjay Ghemawat, Geoffrey Irving, Michael Isard, et~al.
\newblock Tensorflow: a system for large-scale machine learning.
\newblock In {\em OSDI}, volume~16, pages 265--283, 2016.

\bibitem{agrawal2015learning}
Pulkit Agrawal, Joao Carreira, and Jitendra Malik.
\newblock Learning to see by moving.
\newblock In {\em ICCV}, pages 37--45, 2015.

\bibitem{bengio2013representation}
Yoshua Bengio, Aaron Courville, and Pascal Vincent.
\newblock Representation learning: A review and new perspectives.
\newblock {\em IEEE transactions on pattern analysis and machine intelligence},
  35(8):1798--1828, 2013.

\bibitem{bertini2012multi}
Marco Bertini, Alberto Del~Bimbo, and Lorenzo Seidenari.
\newblock Multi-scale and real-time non-parametric approach for anomaly
  detection and localization.
\newblock {\em Computer Vision and Image Understanding}, 116(3):320--329, 2012.

\bibitem{boiman2007detecting}
Oren Boiman and Michal Irani.
\newblock Detecting irregularities in images and in video.
\newblock {\em International journal of computer vision}, 74(1):17--31, 2007.

\bibitem{brattoli2017lstm}
Biagio Brattoli, Uta B{\"u}chler, Anna-Sophia Wahl, Martin~E Schwab, and
  Bj{\"o}rn Ommer.
\newblock Lstm self-supervision for detailed behavior analysis.
\newblock In {\em CVPR}, volume~2, 2017.

\bibitem{caron2018deep}
Mathilde Caron, Piotr Bojanowski, Armand Joulin, and Matthijs Douze.
\newblock Deep clustering for unsupervised learning of visual features.
\newblock In {\em ECCV}, 2018.

\bibitem{chan2008ucsd}
Antoni Chan and Nuno Vasconcelos.
\newblock Ucsd pedestrian dataset.
\newblock {\em IEEE Trans. on Pattern Analysis and Machine Intelligence
  (TPAMI)}, 30(5):909--926, 2008.

\bibitem{chu2017stacked}
Wenqing Chu and Deng Cai.
\newblock Stacked similarity-aware autoencoders.
\newblock In {\em Proceedings of the 26th International Joint Conference on
  Artificial Intelligence}, pages 1561--1567. AAAI Press, 2017.

\bibitem{cover1967nearest}
Thomas Cover and Peter Hart.
\newblock Nearest neighbor pattern classification.
\newblock {\em IEEE transactions on information theory}, 13(1):21--27, 1967.

\bibitem{deng2009imagenet}
Jia Deng, Wei Dong, Richard Socher, Li-Jia Li, Kai Li, and Li Fei-Fei.
\newblock Imagenet: A large-scale hierarchical image database.
\newblock In {\em CVPR}, pages 248--255. Ieee, 2009.

\bibitem{doersch2015unsupervised}
Carl Doersch, Abhinav Gupta, and Alexei~A Efros.
\newblock Unsupervised visual representation learning by context prediction.
\newblock In {\em CVPR}, pages 1422--1430, 2015.

\bibitem{donahue2016adversarial}
Jeff Donahue, Philipp Kr{\"a}henb{\"u}hl, and Trevor Darrell.
\newblock Adversarial feature learning.
\newblock {\em arXiv preprint arXiv:1605.09782}, 2016.

\bibitem{everingham2015pascal}
Mark Everingham, SM~Ali Eslami, Luc Van~Gool, Christopher~KI Williams, John
  Winn, and Andrew Zisserman.
\newblock The pascal visual object classes challenge: A retrospective.
\newblock {\em International journal of computer vision}, 111(1):98--136, 2015.

\bibitem{everingham2010pascal}
Mark Everingham, Luc Van~Gool, Christopher~KI Williams, John Winn, and Andrew
  Zisserman.
\newblock The pascal visual object classes (voc) challenge.
\newblock {\em International journal of computer vision}, 88(2):303--338, 2010.

\bibitem{gidaris2018unsupervised}
Spyros Gidaris, Praveer Singh, and Nikos Komodakis.
\newblock Unsupervised representation learning by predicting image rotations.
\newblock In {\em ICLR}, 2018.

\bibitem{girshick2015fast}
Ross Girshick.
\newblock Fast r-cnn.
\newblock In {\em ICCV}, pages 1440--1448, 2015.

\bibitem{ham2004kernel}
Jihun Ham, Daniel~D Lee, Sebastian Mika, and Bernhard Sch{\"o}lkopf.
\newblock A kernel view of the dimensionality reduction of manifolds.
\newblock In {\em Proceedings of the twenty-first international conference on
  Machine learning}, page~47, 2004.

\bibitem{he2005neighborhood}
Xiaofei He, Deng Cai, Shuicheng Yan, and Hong-Jiang Zhang.
\newblock Neighborhood preserving embedding.
\newblock In {\em ICCV}, volume~2, pages 1208--1213. IEEE, 2005.

\bibitem{hinton2006reducing}
Geoffrey~E Hinton and Ruslan~R Salakhutdinov.
\newblock Reducing the dimensionality of data with neural networks.
\newblock {\em science}, 313(5786):504--507, 2006.

\bibitem{jayaraman2015learning}
Dinesh Jayaraman and Kristen Grauman.
\newblock Learning image representations tied to ego-motion.
\newblock In {\em ICCV}, pages 1413--1421, 2015.

\bibitem{jenni2018self}
Simon Jenni and Paolo Favaro.
\newblock Self-supervised feature learning by learning to spot artifacts.
\newblock In {\em CVPR}, pages 2733--2742, 2018.

\bibitem{karpathy2014large}
Andrej Karpathy, George Toderici, Sanketh Shetty, Thomas Leung, Rahul
  Sukthankar, and Li Fei-Fei.
\newblock Large-scale video classification with convolutional neural networks.
\newblock In {\em CVPR}, pages 1725--1732, 2014.

\bibitem{kim2009observe}
Jaechul Kim and Kristen Grauman.
\newblock Observe locally, infer globally: a space-time mrf for detecting
  abnormal activities with incremental updates.
\newblock In {\em CVPR}, pages 2921--2928, 2009.

\bibitem{krahenbuhl2015data}
Philipp Kr{\"a}henb{\"u}hl, Carl Doersch, Jeff Donahue, and Trevor Darrell.
\newblock Data-dependent initializations of convolutional neural networks.
\newblock {\em arXiv preprint arXiv:1511.06856}, 2015.

\bibitem{krizhevsky2012imagenet}
Alex Krizhevsky, Ilya Sutskever, and Geoffrey~E Hinton.
\newblock Imagenet classification with deep convolutional neural networks.
\newblock In {\em Advances in neural information processing systems}, pages
  1097--1105, 2012.

\bibitem{larsson2016learning}
Gustav Larsson, Michael Maire, and Gregory Shakhnarovich.
\newblock Learning representations for automatic colorization.
\newblock In {\em ECCV}, pages 577--593. Springer, 2016.

\bibitem{lecun1998mnist}
Yann LeCun, Corinna Cortes, and Christopher Burges.
\newblock Mnist dataset.
\newblock {\em URL http://yann. lecun. com/exdb/mnist}, 1998.

\bibitem{lee2017unsupervised}
Hsin-Ying Lee, Jia-Bin Huang, Maneesh Singh, and Ming-Hsuan Yang.
\newblock Unsupervised representation learning by sorting sequences.
\newblock In {\em ICCV}, pages 667--676. IEEE, 2017.

\bibitem{li2018unsupervised}
Minxian Li, Xiatian Zhu, and Shaogang Gong.
\newblock Unsupervised person re-identification by deep learning tracklet
  association.
\newblock In {\em ECCV}, 2018.

\bibitem{li2014anomaly}
Weixin Li, Vijay Mahadevan, and Nuno Vasconcelos.
\newblock Anomaly detection and localization in crowded scenes.
\newblock {\em IEEE transactions on pattern analysis and machine intelligence},
  36(1):18--32, 2014.

\bibitem{long2015fully}
Jonathan Long, Evan Shelhamer, and Trevor Darrell.
\newblock Fully convolutional networks for semantic segmentation.
\newblock In {\em CVPR}, pages 3431--3440, 2015.

\bibitem{mahadevan2010anomaly}
Vijay Mahadevan, Weixin Li, Viral Bhalodia, and Nuno Vasconcelos.
\newblock Anomaly detection in crowded scenes.
\newblock In {\em CVPR}, pages 1975--1981, 2010.

\bibitem{makhzani2015adversarial}
Alireza Makhzani, Jonathon Shlens, Navdeep Jaitly, Ian Goodfellow, and Brendan
  Frey.
\newblock Adversarial autoencoders.
\newblock {\em arXiv preprint arXiv:1511.05644}, 2015.

\bibitem{meng2017magnet}
Dongyu Meng and Hao Chen.
\newblock Magnet: a two-pronged defense against adversarial examples.
\newblock In {\em Proceedings of the 2017 ACM SIGSAC Conference on Computer and
  Communications Security}, pages 135--147. ACM, 2017.

\bibitem{misra2016shuffle}
Ishan Misra, C~Lawrence Zitnick, and Martial Hebert.
\newblock Shuffle and learn: unsupervised learning using temporal order
  verification.
\newblock In {\em ECCV}, pages 527--544. Springer, 2016.

\bibitem{noroozi2016unsupervised}
Mehdi Noroozi and Paolo Favaro.
\newblock Unsupervised learning of visual representations by solving jigsaw
  puzzles.
\newblock In {\em European Conference on Computer Vision}, pages 69--84.
  Springer, 2016.

\bibitem{noroozi2017representation}
Mehdi Noroozi, Hamed Pirsiavash, and Paolo Favaro.
\newblock Representation learning by learning to count.

\bibitem{owens2016ambient}
Andrew Owens, Jiajun Wu, Josh~H McDermott, William~T Freeman, and Antonio
  Torralba.
\newblock Ambient sound provides supervision for visual learning.
\newblock In {\em ECCV}, pages 801--816. Springer, 2016.

\bibitem{papernot2016cleverhans}
Nicolas Papernot, Nicholas Carlini, Ian Goodfellow, Reuben Feinman, Fartash
  Faghri, Alexander Matyasko, Karen Hambardzumyan, Yi-Lin Juang, Alexey
  Kurakin, Ryan Sheatsley, et~al.
\newblock cleverhans v2. 0.0: an adversarial machine learning library.
\newblock {\em preprint arXiv:1610.00768}, 2016.

\bibitem{pathak2017learning}
Deepak Pathak, Ross~B Girshick, Piotr Doll{\'a}r, Trevor Darrell, and Bharath
  Hariharan.
\newblock Learning features by watching objects move.
\newblock In {\em CVPR}, volume~1, page~7, 2017.

\bibitem{pathakCVPR16context}
Deepak Pathak, Philipp Kr\"ahenb\"uhl, Jeff Donahue, Trevor Darrell, and Alexei
  Efros.
\newblock Context encoders: Feature learning by inpainting.
\newblock In {\em CVPR}, 2016.

\bibitem{pathak2016context}
Deepak Pathak, Philipp Krahenbuhl, Jeff Donahue, Trevor Darrell, and Alexei~A
  Efros.
\newblock Context encoders: Feature learning by inpainting.
\newblock In {\em CVPR}, pages 2536--2544, 2016.

\bibitem{ravanbakhsh2017abnormal}
Mahdyar Ravanbakhsh, Moin Nabi, Enver Sangineto, Lucio Marcenaro, Carlo
  Regazzoni, and Nicu Sebe.
\newblock Abnormal event detection in videos using generative adversarial nets.
\newblock In {\em IEEE International Conference on Image Processing (ICIP)},
  2017.

\bibitem{ravanbakhsh2017training}
Mahdyar Ravanbakhsh, Enver Sangineto, Moin Nabi, and Nicu Sebe.
\newblock Training adversarial discriminators for cross-channel abnormal event
  detection in crowds.
\newblock {\em arXiv preprint arXiv:1706.07680}, 2017.

\bibitem{sabokrou2016video}
M Sabokrou, M Fathy, and M Hoseini.
\newblock Video anomaly detection and localisation based on the sparsity and
  reconstruction error of auto-encoder.
\newblock {\em Electronics Letters}, 52(13):1122--1124.

\bibitem{sabokrou2015real}
Mohammad Sabokrou, Mahmood Fathy, Mojtaba Hoseini, and Reinhard Klette.
\newblock Real-time anomaly detection and localization in crowded scenes.
\newblock In {\em CVPR Workshops}, pages 56--62, 2015.

\bibitem{sabokrou2017fast}
Mohammad Sabokrou, Mahmood Fathy, Zahra Moayed, and Reinhard Klette.
\newblock Fast and accurate detection and localization of abnormal behavior in
  crowded scenes.
\newblock {\em Machine Vision and Applications}, 28(8):965--985, 2017.

\bibitem{sabokrou2017deep}
Mohammad Sabokrou, Mohsen Fayyaz, Mahmood Fathy, and Reinhard Klette.
\newblock Deep-cascade: Cascading 3d deep neural networks for fast anomaly
  detection and localization in crowded scenes.
\newblock {\em IEEE Transactions on Image Processing}, 26(4):1992--2004, 2017.

\bibitem{sabokrou2018deep}
Mohammad Sabokrou, Mohsen Fayyaz, Mahmood Fathy, Zahra Moayed, and Reinhard
  Klette.
\newblock Deep-anomaly: Fully convolutional neural network for fast anomaly
  detection in crowded scenes.
\newblock {\em Computer Vision and Image Understanding}, 172:88--97, 2018.

\bibitem{sabokrou2018adversarially}
Mohammad Sabokrou, Mohammad Khalooei, Mahmood Fathy, and Ehsan Adeli.
\newblock Adversarially learned one-class classifier for novelty detection.
\newblock In {\em CVPR}, pages 3379--3388, 2018.

\bibitem{sabokrou2018avid}
Mohammad Sabokrou, Masoud Pourreza, Mohsen Fayyaz, Rahim Entezari, Mahmood
  Fathy, J{\"u}rgen Gall, and Ehsan Adeli.
\newblock Avid: Adversarial visual irregularity detection.
\newblock {\em ACCV}, 2018.

\bibitem{samangouei2018defense}
Pouya Samangouei, Maya Kabkab, and Rama Chellappa.
\newblock Defense-gan: Protecting classifiers against adversarial attacks using
  generative models.
\newblock {\em ICLR}, 2018.

\bibitem{sharifipour2019unsupervised}
Sasan Sharifipour, Hossein Fayyazi, Mohammad Sabokrou, and Ehsan Adeli.
\newblock Unsupervised feature ranking and selection based on autoencoders.
\newblock In {\em ICASSP}, pages 3172--3176. IEEE, 2019.

\bibitem{shawe2004kernel}
John Shawe-Taylor, Nello Cristianini, et~al.
\newblock {\em Kernel methods for pattern analysis}.
\newblock Cambridge university press, 2004.

\bibitem{vapnik1995support}
Vladimir Vapnik, Isabel Guyon, and Trevor Hastie.
\newblock Support vector machines.
\newblock {\em Mach. Learn}, 20(3):273--297, 1995.

\bibitem{vincent2008extracting}
Pascal Vincent, Hugo Larochelle, Yoshua Bengio, and Pierre-Antoine Manzagol.
\newblock Extracting and composing robust features with denoising autoencoders.
\newblock In {\em Proceedings of the 25th international conference on Machine
  learning}, pages 1096--1103. ACM, 2008.

\bibitem{wang2017feature}
Shuyang Wang, Zhengming Ding, and Yun Fu.
\newblock Feature selection guided auto-encoder.
\newblock In {\em AAAI}, pages 2725--2731, 2017.

\bibitem{wang2015unsupervised}
Xiaolong Wang and Abhinav Gupta.
\newblock Unsupervised learning of visual representations using videos.
\newblock In {\em ICCV}, pages 2794--2802, 2015.

\bibitem{wang2004image}
Zhou Wang, Alan~C Bovik, Hamid~R Sheikh, and Eero~P Simoncelli.
\newblock Image quality assessment: from error visibility to structural
  similarity.
\newblock {\em IEEE transactions on image processing}, 13(4):600--612, 2004.

\bibitem{wu2018unsupervised}
Zhirong Wu, Yuanjun Xiong, X~Yu Stella, and Dahua Lin.
\newblock Unsupervised feature learning via non-parametric instance
  discrimination.
\newblock In {\em CVPR}, pages 3733--3742, 2018.

\bibitem{xu2015learning}
Dan Xu, Elisa Ricci, Yan Yan, Jingkuan Song, and Nicu Sebe.
\newblock Learning deep representations of appearance and motion for anomalous
  event detection.
\newblock In {\em BMVC}, 2015.

\bibitem{zhang2016colorful}
Richard Zhang, Phillip Isola, and Alexei~A Efros.
\newblock Colorful image colorization.
\newblock In {\em ECCV}, pages 649--666. Springer, 2016.

\bibitem{zhang2017split}
Richard Zhang, Phillip Isola, and Alexei~A Efros.
\newblock Split-brain autoencoders: Unsupervised learning by cross-channel
  prediction.
\newblock In {\em CVPR}, volume~1, page~5, 2017.

\bibitem{zhao2019variational}
Qingyu Zhao, Nicolas Honnorat, Ehsan Adeli, Adolf Pfefferbaum, Edith~V
  Sullivan, and Kilian~M Pohl.
\newblock Variational autoencoder with truncated mixture of gaussians for
  functional connectivity analysis.
\newblock In {\em International Conference on Information Processing in Medical
  Imaging}, pages 867--879. Springer, 2019.

\end{thebibliography}
}

\end{document}